\documentclass[runningheads]{llncs}

\usepackage{eccv}

\usepackage{eccvabbrv}

\usepackage{graphicx}
\usepackage{booktabs}

\usepackage[accsupp]{axessibility}  %

\usepackage[pagebackref,breaklinks,colorlinks,citecolor=eccvblue]{hyperref}

\usepackage{orcidlink}

\usepackage{rotating}
\usepackage[rightcaption]{sidecap}
\usepackage{gensymb}
\usepackage{relsize}
\usepackage{makecell}

\sidecaptionvpos{figure}{c}

\definecolor{turquoise}{cmyk}{0.65,0,0.1,0.3}
\definecolor{purple}{rgb}{0.65,0,0.65}
\definecolor{dark_green}{rgb}{0, 0.5, 0}
\definecolor{orange}{rgb}{0.8, 0.6, 0.2}
\definecolor{red}{rgb}{0.8, 0.2, 0.2}
\definecolor{darkred}{rgb}{0.6, 0.1, 0.05}
\definecolor{blueish}{rgb}{0.3, 0.3, .6}
\definecolor{light_gray}{rgb}{0.7, 0.7, .7}
\definecolor{pink}{rgb}{1, 0, 1}
\definecolor{greyblue}{rgb}{0.25, 0.25, 1}
\definecolor{awesome}{rgb}{1.0, 0.13, 0.32}

\definecolor{figred}{rgb}{0.9, 0.1, 0.1}
\definecolor{figgreen}{rgb}{0.1, 0.7, 0.1}
\definecolor{figblue}{rgb}{0.1, 0.1, 0.9}
\definecolor{figmagenta}{rgb}{0.8, 0.1, 0.8}

\begin{document}

\title{Generative Camera Dolly:\\Extreme Monocular Dynamic Novel View Synthesis}

\titlerunning{GCD: Extreme Monocular Dynamic Novel View Synthesis}

\author{Basile Van Hoorick\inst{1} \and
Rundi Wu\inst{1} \and
Ege Ozguroglu\inst{1} \and
Kyle Sargent\inst{2} \and
Ruoshi Liu\inst{1} \and \\
Pavel Tokmakov\inst{3} \and
Achal Dave\inst{3} \and
Changxi Zheng\inst{1} \and
Carl Vondrick\inst{1}
}

\authorrunning{B.~Van Hoorick et al.}

\institute{Columbia University \and
Stanford University \and
Toyota Research Institute\\[0.22cm]
{\larger \href{https://gcd.cs.columbia.edu/}{gcd.cs.columbia.edu}}}

\maketitle

\begin{abstract}
Accurate reconstruction of complex dynamic scenes from just a single viewpoint continues to be a challenging task in computer vision. 
Current dynamic novel view synthesis methods typically require videos from many different camera viewpoints, necessitating careful recording setups, and significantly restricting their utility in the wild as well as in terms of embodied AI applications.
In this paper, we propose \textbf{GCD}, a controllable monocular dynamic view synthesis pipeline that leverages large-scale diffusion priors to, given a video of any scene, generate a synchronous video from any other chosen perspective, conditioned on a set of relative camera pose parameters.
Our model does not require depth as input, and does not explicitly model 3D scene geometry, instead performing end-to-end video-to-video translation in order to achieve its goal efficiently.
Despite being trained on synthetic multi-view video data only, zero-shot real-world generalization experiments show promising results in multiple domains, including robotics, object permanence, and driving environments.
We believe our framework can potentially unlock powerful applications in rich dynamic scene understanding, perception for robotics, and interactive 3D video viewing experiences for virtual reality.
\end{abstract}

\begin{figure}[t]
  \centering
  \includegraphics[width=\linewidth]{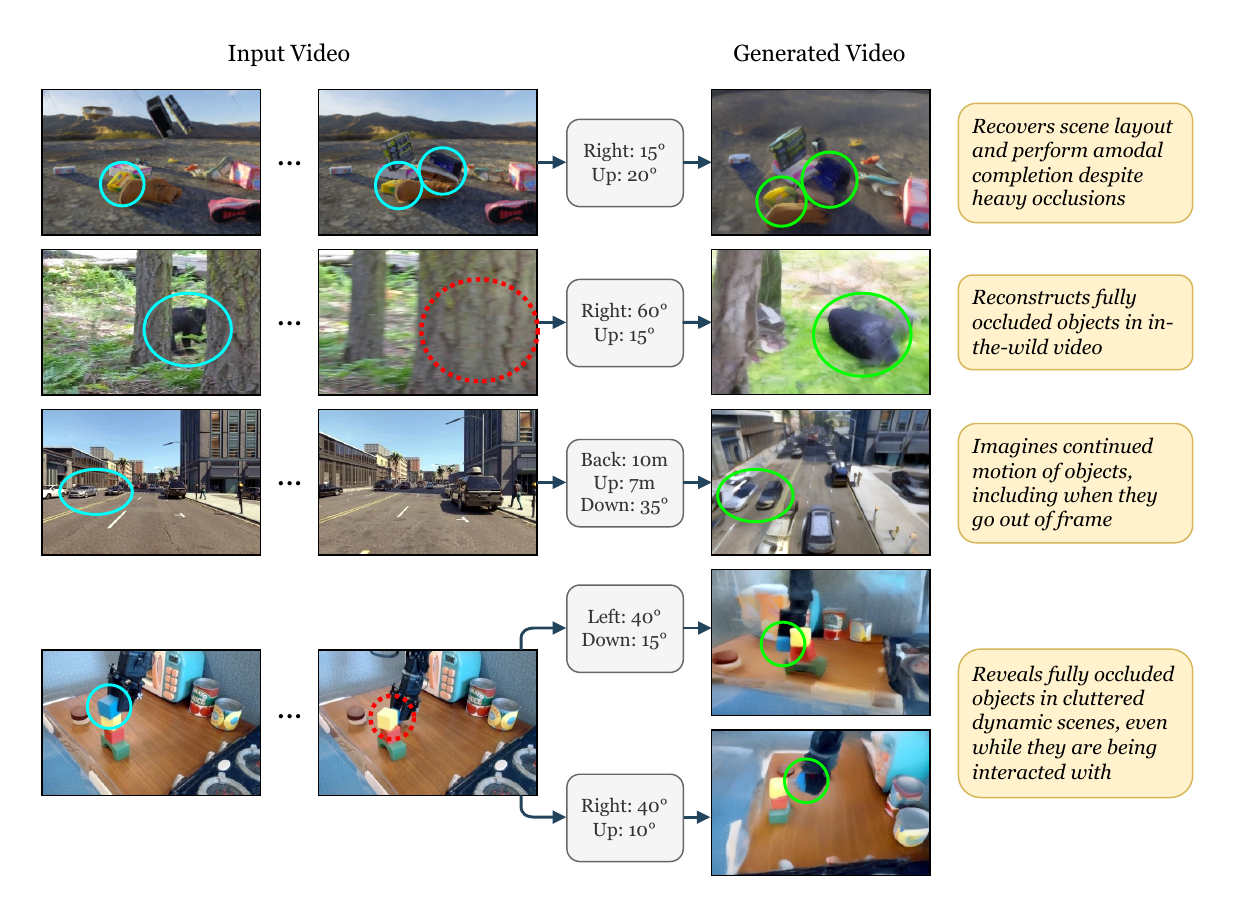}
  \vspace{-0.05in}
  \caption{
  \textbf{Spatial video translation of dynamic scenes.}
  Given a single RGB video, we propose a method that is capable of imagining what that scene would look like from another viewpoint.
  Even for extreme camera transformations with large angles,
  our approach synthesizes videos with rich visual details that are consistent with the input, demonstrating advanced spatiotemporal reasoning capabilities.
  \vspace{-0.1in}
  }
  \label{fig:teaser}
\end{figure}

\section{Introduction}
\label{sec:intro}

Video generation has made tremendous progress in recent years. Results from Sora~\cite{videoworldsimulators2024}, OpenAI's recently released text-to-video generation model, have shown that generating a high-quality video as long as one minute is possible. Following the scaling curve, video models will most likely continue to improve in many aspects. However, one essential capability is still missing from these video models to be useful for many downstream applications -- the ability to generate the same dynamic scene from an arbitrary camera perspective based on an existing video. %

In this paper, we aim to tackle the problem of \emph{dynamic novel view synthesis} (DVS) -- given a video of a dynamic scene, we aim to generate a video of the same scene from another viewpoint. Once we develop a solution for this problem, we can leverage it for several impactful use cases, such as generating novel views of a live street scenario based on cameras mounted on an autonomous vehicle; seeing a cluttered environment from a different viewpoint while a robot is performing dexterous manipulations; enabling geometrically consistent video passthrough for mixed reality~\cite{xiao2022neuralpassthrough}; and immersively reliving videos recorded in the past from different viewing angles.

However, this task is naturally extremely ill-posed and challenging. While yielding promising results, prior works typically addressed it by assuming either that contemporaneous multi-viewpoint video is available~\cite{pumarola2021dnerf,wang2022fourier,xie2022neural,wu20234dgaussians,luiten2023dynamic}, and/or by imposing that the relative camera viewpoint changes must be small (\ie limited to just a handful of degrees)~\cite{li2023dynibar,wang2024diffusion}. These restrictions make them vastly insufficient for the aforementioned applications, which require in-the-wild novel view synthesis pipelines with dramatic camera viewpoint changes.

Free-viewpoint synthesis from a single video requires prior knowledge because it is highly under-constrained. Modern video generative models, such as Stable Video Diffusion~\cite{blattmann2023stable}, have learned rich priors for real-world dynamics, 3D geometry, and camera motions, as they are trained on hundreds of millions of video clips from the Internet. In this work, we propose an approach to capitalize on these rich representations for the task of DVS. We curate pairs of videos of dynamic scenes from simulation as training data, and apply them to steer a pretrained video generative model towards the desired behavior by means of finetuning.

Qualitative and quantitative results demonstrate that our model achieves state-of-the-art results on the task of monocular DVS, and generalizes effectively to various out-of-distribution scenes, including real-world driving videos, robot manipulation scenes, and other in-the-wild videos with heavy occlusion patterns, as shown in Figure~\ref{fig:teaser}. Much like a camera dolly in film-making~\cite{dolly}, our approach essentially conceives a virtual camera that can move around with up to six degrees of freedom, reveal significant portions of the scene that are otherwise unseen, reconstruct hidden objects behind occlusions, all within complex dynamic scenes, even when the contents are moving.

Our core contribution is the design and evaluation of a framework, \emph{Generative Camera Dolly} (GCD), for learning to generate videos from novel viewpoints of a dynamic scene, using an end-to-end video-to-video neural network.
Section~\ref{sec:rel} provides a brief overview of related work.
Section~\ref{sec:method} introduces the approach including the model architecture, and a description of how to achieve precise camera control within the video diffusion model.
Section~\ref{sec:data} discusses training data, benchmarks, and task details.
Section~\ref{sec:traject} investigates important hyperparameter decisions with regard to the conceptual implementation of camera control.
Section~\ref{sec:exp} provides both quantitative and qualitative evaluation of the system as well as several examples of our model generalizing to out-of-distribution data.
We believe the ability to perform free-viewpoint video synthesis for a dynamic scene from one video will have a significant impact on 3D/4D computer vision research, as well as other related areas, including content creation, AR/VR, and robotics.

\section{Related Work}
\label{sec:rel}

\paragraph{Dynamic scene reconstruction.}
The landscape of dynamic scene novel view synthesis has been primarily dominated by techniques that rely on multiple synchronized (\ie contemporaneous) input videos~\cite{bansal20204d,bemana2020x,li2022neural,broxton2020immersive,zhang2021editable,pumarola2021dnerf,wang2022fourier,xie2022neural}, which limits their practical usage in real-world scenarios.
The emergence of Neural Radiance Fields (NeRF)~\cite{mildenhall2021nerf} has catalyzed a revolution in dynamic view synthesis, presenting state-of-the-art results in this domain~\cite{xian2021space,tretschk2021non,li2021neural,pumarola2021d,du2021neural,park2021nerfies,park2021hypernerf}.
Most such methods represent scenes through time-evolving NeRFs, for handling complicated, dynamic 3D scene motions in casual videos, for example in neural scene flow fields~\cite{li2021neural,xian2021space,wang2021neural,gao2021dynamic,gao2022monocular,cao2023hexplane}.

A notable trend in recent advancements involves the synthesis of novel views from a single camera perspective~\cite{yoon2020novel,gao2022monocular,li2023dynibar,wang2024diffusion}.
DynIBaR adopts a volumetric image-based rendering framework that, instead of encoding and compressing the entire scene within a single representation (for example an MLP), aggregates features from nearby views in a camera motion-aware manner, which enables synthesizing novel views for long videos with uncontrolled camera paths~\cite{li2023dynibar}.
DpDy leverages an image-based diffusion model to iteratively distill knowledge coming from diffusion priors into a hybrid 4D representation, consisting of a static and dynamic NeRF~\cite{wang2024diffusion}.

It is worth noting that essentially all aforementioned methods optimize \emph{per-scene} representations independently of each other. Therefore, they are (1) largely unable to share any knowledge between different reconstructions, such as to generalize to unseen environments; and (2) largely unable to infer or extrapolate from incomplete observations, such as to recover fully occluded regions. Moreover, failure modes are often observed when the monocular input video lacks \emph{effective} multi-view cues, for example as enabled implicitly thanks to a moving, especially a fast-moving, camera~\cite{gao2022monocular}. Exceptions include~\cite{van2022revealing},
where dynamic scene completion is performed through a conditional neural field based on a single, static RGB-D input video.

\paragraph{Video diffusion models.}
Recent work has rapidly improved the state of video generation models. Most generative models focus on diffusion-based approaches~\cite{singer2022make,ho2022imagen,Ho2022VideoDM,blattmann2023align,ge2023preserve,blattmann2023stable}, though important exceptions exist, particularly with autoregressive training~\cite{yu2023language,weissenborn2019scaling}.
Following recent work which shows image-based diffusion models can be re-purposed for computer vision tasks including monocular depth estimation~\cite{saxena2023monocular}, 3D reconstruction~\cite{liu2023zero} and amodal segmentation~\cite{ozguroglu2024pix2gestalt}, our work adopts a public video diffusion model for dynamic view synthesis.
We rely on Stable Video Diffusion~\cite{blattmann2023stable} as it generates high quality videos, and provides a public image-to-video model checkpoint with code, although our framework can generalize to any video generation approach.

\paragraph{3D and 4D generation.}
Most of the works enabling successful 3D generation via generative models hence rely on channelling the representational power of 2D diffusion models towards a single 3D representation that is iteratively optimized over time, for example through score distillation~\cite{poole2022dreamfusion}.
This \emph{multiview 2D-to-3D} paradigm is exemplified by many text-to-3D and image-to-3D works~\cite{liu2023zero,poole2022dreamfusion,lin2023magic3d,wang2023score,wang2023prolificdreamer,chen2023fantasia3d,haque2023instruct,po2023compositional,zhang2023scenewiz3d,hollein2023text2room,wu2023reconfusion,voleti2024sv3d}.

Emphasizing the temporal component, text-to-4D and image-to-4D papers have begun appearing as well, although the results currently remain mostly limited to animations of single objects or animals~\cite{singer2023text,bahmani20234d,ling2023align,zhao2023animate124}.
Video-to-4D, which is likely harder because
every frame of the observation must be respected, has remained less explored so far. In~\cite{van2022revealing}, a video-to-4D scene reconstruction task and framework is proposed, although the model requires depth input, and only works in narrow domains as it is trained from scratch.

\paragraph{Object permanence and amodal completion.}
The problem of reasoning about the invisible parts of a scene has been studied extensively in the literature, but so far almost exclusively from an object-centric perspective. For example, in the image world, amodal completion~\cite{ehsani2018segan,zhan2020self,ozguroglu2024pix2gestalt} studies the problem of reconstructing the occluded parts of an object based on its visible parts and the scene context. However, these methods are naturally restricted to partial occlusions. In contrast, for videos, some object tracking methods can capitalize on the temporal context to successfully reason about the location~\cite{tokmakov2021learning,tokmakov2022object, shamsian2020learning} or even shape~\cite{van2023tracking} of fully occluded instances. 

While abstracting the full complexity of a dynamic scene into a compact set of objects allows these methods to be relatively data- and compute-efficient, it also limits their applicability. In this work, we propose a more general approach that is capable of revealing  any parts of a scene, together with their dynamics, similar to~\cite{van2022revealing}.
This includes not only occluded objects, but also `stuff' regions~\cite{caesar2018coco},
such as natural or man-made surfaces, liquids, and so on.

We note that at least one concurrent work also tackles dynamic view synthesis: in Exo2Ego~\cite{luo2024put}, authors propose a framework that translates third-person (exocentric) to first-person (egocentric) videos on a per-frame basis, incorporating priors for hand-object interactions and focusing primarily on those scenarios.

\begin{figure}[t]
  \centering
  \includegraphics[width=\linewidth]{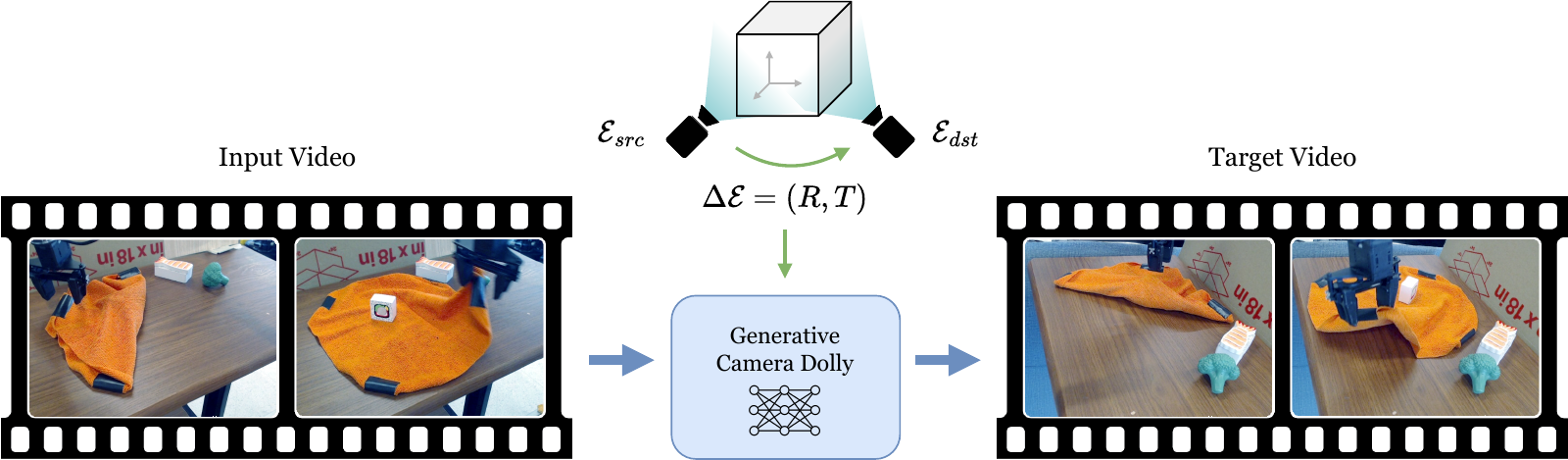}
  \vspace{-0.05in}
  \caption{
  \textbf{Method.}
  Our model, GCD, is an end-to-end video translation pipeline that maps an input video from any viewpoint into an output video from any other perspective, with the objective of respecting all objects and dynamics occurring within the observed dynamic scene, and faithfully reconstructing the corresponding visual details from this novel viewpoint. The relative camera extrinsics matrix $\Delta \mathcal{E}$ guides the relationship between the two camera poses.
  \vspace{-0.1in}
  }
  \label{fig:method}
\end{figure}

\section{Approach}
\label{sec:method}

First, we formally introduce the task of monocular dynamic novel view synthesis from unconstrained video input. Let $\boldsymbol x \in \mathbb{R}^{T \times H \times W \times 3}$ be RGB frames captured from a single camera perspective, that encode the visual observation of a dynamic scene of interest.
We denote its associated camera extrinsics matrix as $\mathcal{E}_{src} \in \mathbb{R}^{T \times 4 \times 4}$, and define $\mathcal{E}_{dst} \in \mathbb{R}^{T \times 4 \times 4}$ to be the desired target camera extrinsics matrix over time. Our model $f$ is then tasked with predicting a video $\boldsymbol y \in \mathbb{R}^{T \times H \times W \times 3}$, that plausibly depicts the same dynamic scene from the specified new viewpoint. For simplicity, and without loss of generality, we assume that (1) the output video is temporally synchronized with the input, and (2) the camera intrinsics matrix $\mathcal{K} \in \mathcal{R}^{3 \times 3}$ stays constant over time as well as across pose changes; notably, the virtual camera for $\boldsymbol y$ assumes the same focal length as the actual camera does for $\boldsymbol x$.

Since novel view synthesis is an inherently under-constrained, challenging problem, our approach will use existing large-scale video generative models.
Diffusion models have been shown to excel at image-to-3D tasks~\cite{liu2023zero,shi2023mvdream,long2023wonder3d,wu2023reconfusion}, justifying our attempt to perform video-to-4D. Moreover, they have shown remarkable zero-shot abilities in generating realistic, diverse videos from user-given text descriptions and/or initial frames~\cite{Ho2022VideoDM,blattmann2023stable,bar2024lumiere}. However, they are typically not trained to accept video as a conditioning signal, and fine-grained control over camera transformations is also not available by default. To overcome these obstacles, we must make a few architectural changes.

\subsection{Camera viewpoint control}

Given a single RGB video $\boldsymbol x$ of a dynamic scene, our goal is to synthesize another video $\boldsymbol y$ of the scene from a different viewpoint.  
Since large-scale video diffusion models have been trained on hyper-scale data, their support of the natural video distribution most likely covers a wide range of realistic scenes and viewpoints.
To this end, given a dataset of paired videos and their \emph{relative} camera extrinsics $\Delta \mathcal{E} = \{ \mathcal{E}_{src,t}^{-1} \cdot \mathcal{E}_{dst,t} \}_{t=0}^{T-1} \in \mathbb{R}^{T \times 4 \times 4}$ over time, we teach a latent diffusion model $f$ to learn controls over camera parameters within any video $\boldsymbol x$:
\begin{equation}
    \boldsymbol y = f \left( \boldsymbol x, \Delta \mathcal{E} \right)
\end{equation}
Specifically, we modify Stable Video Diffusion (SVD) to accept a new form of \emph{micro-conditioning}, a term coined in~\cite{blattmann2023stable}, which is designed for the purpose of communicating low-dimensional metadata (such as the desired frame rate of the output video, and the amount of optical flow) to the network. We decompose $\Delta \mathcal{E}_t \in \mathrm{SE}(3)$ into a series of camera rotation matrices $R_t \in \mathrm{SO}(3)$ and translation matrices $T_t \in \mathbb{R}^3$ over time, project this flattened information through an MLP $m$, and add the resulting embedding to the feature vectors at various convolutional layers placed throughout the network, similarly to the concurrent work SV3D~\cite{voleti2024sv3d}.
The diffusion timestep, FPS, and motion strength are also passed to the network this way.  %
To preserve the existing priors of SVD as much as possible, we initialize the network weights based on the publicly available image-to-video model checkpoint. The new embedder $m$ that processes $\{(R_t,T_t)\}$ is randomly initialized with default parameters.
After training the network end-to-end to tackle this new task, the resulting model is capable of imagining unseen videos from any chosen perspective, as illustrated in Figure~\ref{fig:method} (high-level) and Figure~\ref{fig:arch} (detail).

\subsection{Video conditioning}

To accurately perform dynamic view synthesis,
both low-level perception (to analyze the visible geometry, shapes, appearance, etc.) and high-level understanding (to infer the occluded regions, based on world knowledge as well as other observed frames) of the input video is required. We adopt the same hybrid conditioning mechanism as SVD~\cite{blattmann2023stable}, where the visual signal is processed in two ways. In case of image-to-video, the first stream calculates the CLIP~\cite{radford2021learning} embedding $c(\boldsymbol{x}_0)$ of the incoming image to condition the U-Net $\epsilon$ via cross-attention, and the second stream channel-concatenates the VAE-encoded image $\boldsymbol{x}_0$ with all frames of the video sample $\hat{\boldsymbol{y}}$ that are being denoised. %

We keep this mechanism almost entirely intact when moving from the pretraining to the finetuning stage, but we propose to simply substitute the first frame $\boldsymbol{x}_0$ for the entire input video $\boldsymbol{x}$ from the source viewpoint, such that the conditioning information now becomes a function of time. 
This ensures that our model has the opportunity to watch how the dynamic scene unfolds over time, and hence must learn to respect the dynamics and physics of the objects within.

In architectural terms, the output sample $\hat{\boldsymbol{y}}$ has contemporaneous input frames from $\boldsymbol{x}$ attached to it for every video timestamp $t$, such that at diffusion noise timestep $u$ during inference:
\begin{equation}
    \hat{\boldsymbol{y}}_{u-1} = w \epsilon \left( \hat{\boldsymbol{y}}_u \mathbin\Vert \boldsymbol{x}, \Delta \mathcal{E} \right) - (w-1) \epsilon \left( \hat{\boldsymbol{y}}_u \right),
\end{equation}
where $w \in [1, \infty)$ is the guidance strength for classifier-free guidance~\cite{ho2022classifier}.
The U-Net $\epsilon$ accepts input feature maps of dimensionality $2D \times T \times \frac{H}{F} \times \frac{W}{F}$, where $D$ and $F$ are the VAE embedding size and downsampling factor respectively, and produces output feature maps of dimensionality ${D \times T \times \frac{H}{F} \times \frac{W}{F}}$ that represent a less noisy sample.

Note that the SVD architecture consists of a factorized 3D U-Net that interleaves convolutional, spatial, and temporal blocks, of which the latter two establish correspondences between features across locations (per frame), and across time (per spatial position) respectively. Spatiotemporal attention can consequently take place between all pairs of input and output frames, as well as any pair of regions within both videos. Moreover, there are now $T$ different CLIP embeddings $\{c(\boldsymbol{x}_t)\}$ that appropriately condition the U-Net layers at each matching frame.

\begin{figure}[tb]
  \centering
  \includegraphics[width=\linewidth]{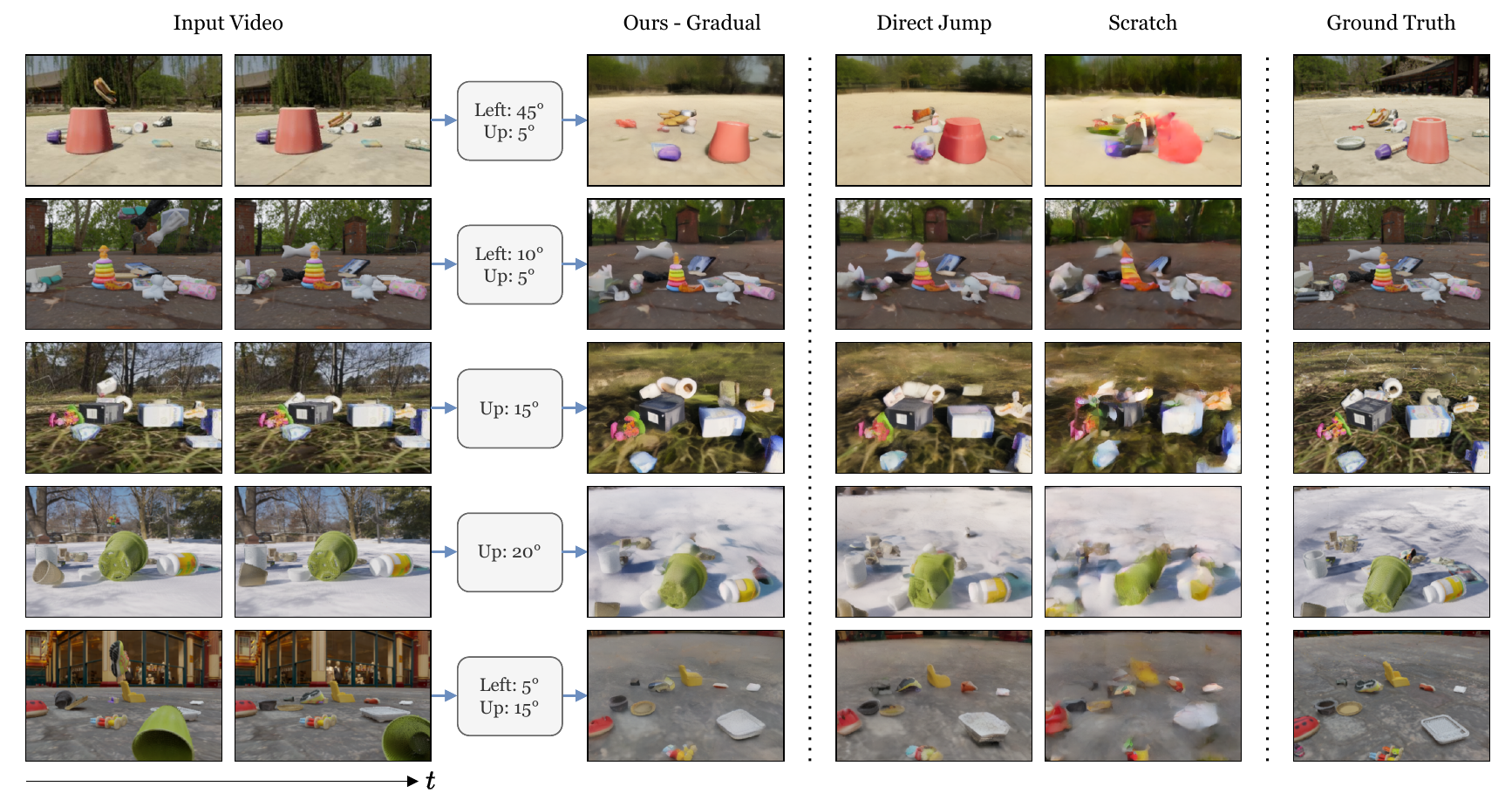}
  \vspace{-0.05in}
  \caption{
  \textbf{Qualitative ablation study results for Kubric-4D.}
  We show inputs, predictions, ablations, and ground truths. The input and output videos both consist of $T=14$ frames, but we show the first and last frame of the input video for conciseness, and only the last frame of the output and target. Whereas the ablations tend to look blurry with incorrect shape and/or appearance characteristics (especially for moving objects), our main model (gradual, max 90\textdegree, finetuned) faithfully reconstructs the scene layout and dynamics from the input video.
  In addition, it often hallucinates plausible backgrounds in unseen regions.
  \vspace{-0.1in}
  }
  \label{fig:kubric_abl}
\end{figure}

\begin{figure}[tb]
  \centering
  \includegraphics[width=\linewidth]{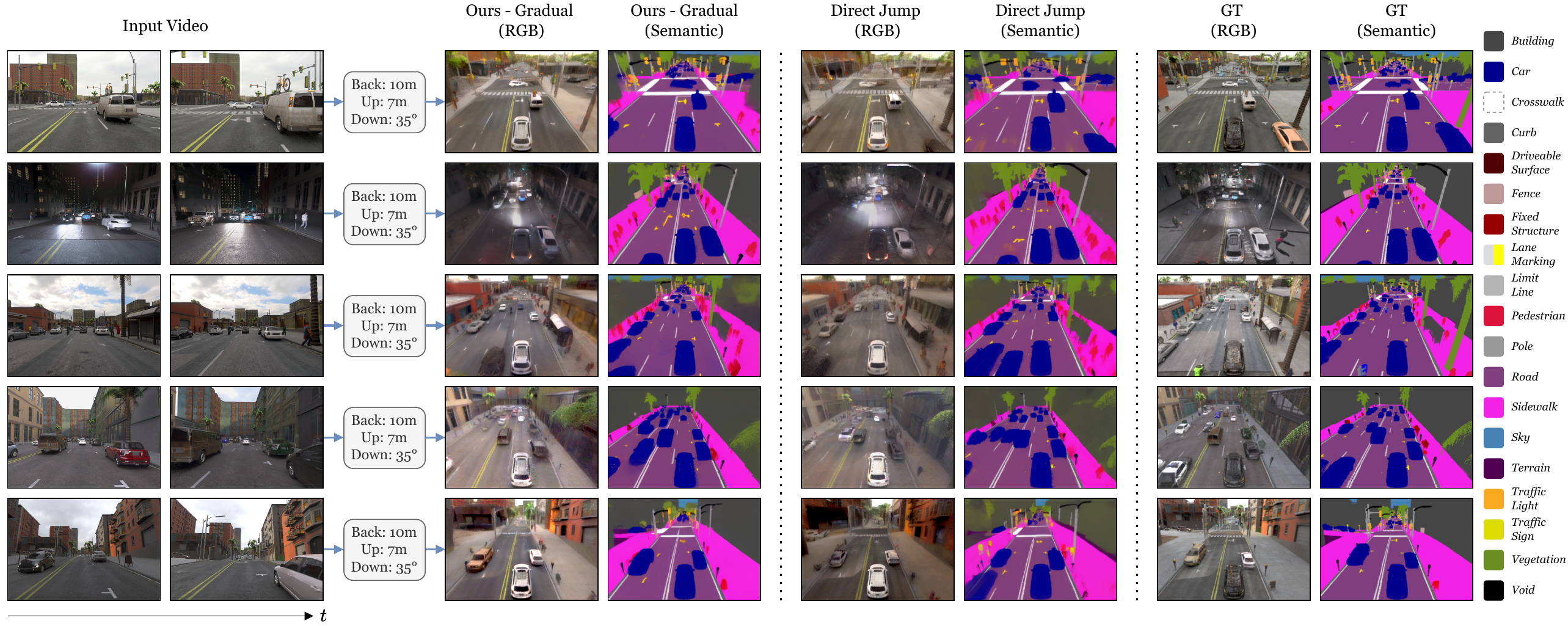}
  \vspace{-0.05in}
  \caption{
  \textbf{Qualitative ablation study results for ParallelDomain-4D.}
  We show inputs, predictions, ablations, and ground truths for both visual and semantic scene completion.
  Our model excels at recovering the top-down viewpoint with high accuracy in both modalities, despite the heavy occlusion patterns that often occur in driving scenes.
  While the \emph{direct} model performs almost as well as the \emph{gradual} one, it tends to introduce slightly more hallucination and discoloration of objects.
  \vspace{-0.1in}
  }
  \label{fig:pardom_abl}
\end{figure}

\section{Datasets}
\label{sec:data}

While the availability of multi-view video data has been growing~\cite{corona2021meva,van2022revealing,sener2022assembly101,grauman2022ego4d,tschernezki2023epic,raistrick2023infinite,zheng2023pointodyssey,grauman2023ego,khirodkar2023ego,raistrick2024infinigen}, it is still relatively sparse compared to conventional image or video datasets.
In order to train and evaluate our model, we require a decent amount of multi-view RGB videos from highly cluttered dynamic scenes. To this end, we contribute two high-quality synthetic datasets, shown in Figures~\ref{fig:kubric_abl},~\ref{fig:pardom_abl}, and~\ref{fig:kubric_sota} and briefly describe them below.

\subsection{Kubric-4D}

We leverage the Kubric~\cite{greff2022kubric} simulator as our data source for generic multi-object interaction videos, carrying a high degree of visual detail and physical realism.
Each scene contains between 7 and 22 randomly sized objects in total, with roughly one-third of them spawned in mid-air at the beginning of the video to encourage sophisticated dynamics. Complicated occlusion patterns arise very frequently, making this dataset highly challenging for accurate novel view synthesis.
We generate 3,000 scenes of 60 frames each, at a frame rate of 24 FPS, with RGB-D data rendered from 16 virtual cameras at a fixed set of poses.

Because the dynamic scene is sufficiently densely covered, we unproject all the pixels from available viewpoints into a merged 3D point cloud per frame. As a form of data augmentation, we then render them into videos from arbitrary viewpoints according to camera trajectories that can be chosen and controllably sampled depending on the exact training configuration.

\subsection{ParallelDomain-4D}

Since rich scene understanding and spatial reasoning skills are paramount for maximizing situational awareness in the context of driving, we employ the state-of-the-art data generation service ParallelDomain to produce complex, highly photorealistic road scenes.
The videos depict driving scenarios covering a wide variety of locations, vehicles, persons, traffic situations, and weather conditions.
Here, we have 1,533 scenes available of 50 frames each, at a frame rate of 10 FPS, with high-quality annotations for multiple modalities (RGB colors, semantic categories, instance IDs, etc.) along with per-pixel ground truth depth rendered from 19 virtual cameras at a fixed set of poses.

In our experiments, we train separate models for RGB view synthesis and semantic view synthesis; the latter demonstrates that the predicted modality need not be the same as the given modality.

Similarly as for Kubric-4D, we perform a unproject-and-reproject routine to turn this multi-view video dataset into a pseudo-4D data source from which we can render videos of the scene from arbitrary camera perspectives over time, within certain pre-defined spatiotemporal bounds.

\subsection{Task details}
\label{sec:task}

In our experiments, without loss of generality, we assume a static input camera pose $\mathcal{P}_{{src}}$,\footnote{This can always be achieved by defining the reference coordinate system to move along with the recording camera.} and pick a target destination pose $\mathcal{P}_{{dst}}$ that we want the output camera to reach at or before the end of the generated video. In general, $\mathcal{P}$ is a \emph{pose description} that can be defined in many ways, for example a set of spherical coordinates that represent the camera position and look-at location, but (1) must allow for convex interpolation (\ie $\alpha \mathcal{P}_1 + (1-\alpha) \mathcal{P}_2$ with $\alpha \in [0,1]$ is valid), and (2) is connected to a valid 6-DoF rigid body transformation $\mathcal{E} \in \mathrm{SE}(3)$ through the function $g$, \ie $\mathcal{E}=g(\mathcal{P})$. %

When training for the task of dynamic view synthesis on Kubric-4D, pairs of input and output poses are randomly sampled within certain spherical coordinate bounds (both in absolute terms and relative to each other),
with the extra condition that they are looking at the center of the 3D scene.
Therefore, at inference time, there are three effective degrees of freedom with regard to camera control, \ie $\mathcal{P} \in \mathbb{R}^3$. %

In case of ParallelDomain-4D, the input video and pose always correspond to the ego vehicle's forward-facing viewpoint, as if a sensor were mounted on the front of the car. The output pose is a fixed top-down viewpoint with the ego vehicle at the bottom center of the video, which enables gaining a much more detailed overview of its surroundings.

\newcommand{\tss}{\textsuperscript}
\newcommand{\tdag}{\textdagger}

\setlength{\tabcolsep}{6pt}

\begin{table}[tb]
  \centering
  \scalebox{0.9}{
  \begin{tabular}{@{}l|rrrrr@{}}
    \toprule
    Variant & \thead{PSNR\\ (all)~$\uparrow$} & \thead{SSIM\\ (all)~$\uparrow$} & \thead{LPIPS\\ (all)~$\downarrow$} & \thead{PSNR\\ (occ.)~$\uparrow$} & \thead{SSIM\\ (occ.)~$\uparrow$} \\
    
    \midrule
    \textbf{Ours} (direct, max 90\textdegree, scratch) & 15.96 & 0.450 & 0.575 & 15.85 & 0.470 \\ %
    \textbf{Ours} (direct, max 180, scratch) & 14.71 & 0.426 & 0.611 & 15.15 & 0.458 \\ %
    \textbf{Ours} (gradual, max 90\textdegree, scratch) & 16.92 & 0.486 & 0.542 & 16.59 & 0.494 \\ %
    \textbf{Ours} (gradual, max 180\textdegree, scratch) & 16.63 & 0.479 & 0.552 & 16.34 & 0.491 \\ %
    
    \midrule
    \textbf{Ours} (direct, max 90\textdegree, finetuned) & 17.23 & 0.494 & 0.507 & 16.69 & 0.492 \\ %
    \textbf{Ours} (direct, max 180\textdegree, finetuned) & 16.65 & 0.471 & 0.529 & 16.18 & 0.470 \\ %
    \textbf{Ours} (gradual, max 90\textdegree, finetuned) & \textbf{17.88} & \textbf{0.521} & \textbf{0.486} & \textbf{17.33} & 0.514 \\ %
    \textbf{Ours} (gradual, max 180\textdegree, finetuned) & 17.81 & \textbf{0.521} & 0.488 & 17.20 & \textbf{0.515} \\ %
  \bottomrule
  \end{tabular}
  }
  \vspace{0.07in}
  \caption{
  \textbf{Ablation study results on Kubric.} We evaluate various versions of our dynamic view synthesis model on only the last frame for fairness, \ie to ensure that the direct and gradual trajectory models are spatially aligned. See Figure~\ref{fig:kubric_abl} for qualitative illustrations.
  \vspace{-0.2in}
  }
  \label{tab:kubric_abl}
\end{table}

\begin{table}[tb]
  \centering
  \scalebox{0.9}{
  \begin{tabular}{@{}l|rrrrr@{}}
    \toprule
    Variant & \thead{PSNR\\ (all)~$\uparrow$} & \thead{SSIM\\ (all)~$\uparrow$} & \thead{LPIPS\\ (all)~$\downarrow$} & \thead{PSNR\\ (occ.)~$\uparrow$} & \thead{SSIM\\ (occ.)~$\uparrow$} \\
    
    \midrule
    \textbf{Ours} (direct, scratch) & 22.49 & 0.622 & 0.487 & 22.62 & 0.653 \\ %
    \textbf{Ours} (gradual, scratch) & 22.73 & 0.632 & 0.467 & 22.76 & 0.664 \\ %
    
    \midrule
    \textbf{Ours} (direct, finetuned) & 23.32 & 0.664 & 0.440 & 23.29 & 0.691 \\ %
    \textbf{Ours} (gradual, finetuned) & \textbf{23.47} & \textbf{0.670} & \textbf{0.425} & \textbf{23.52} & \textbf{0.696} \\ %
    
  \bottomrule
  \end{tabular}
  }
  \vspace{0.07in}
  \caption{
  \textbf{Ablation study results on ParallelDomain in RGB space.}
  We perform visual scene completion, and evaluate various dynamic view synthesis models on only the last frame for fairness, similarly to Table~\ref{tab:kubric_abl}. See Figure~\ref{fig:pardom_abl} for qualitative illustrations.
  \vspace{-0.2in}
  }
  \label{tab:pardom_rgb_abl}
\end{table}

\begin{SCtable}
  \centering
  \scalebox{0.9}{
  \begin{tabular}{@{}l|rr@{}}
    \toprule
    Variant & \thead{mIoU\\ (all)~$\uparrow$} & \thead{mIoU\\ (occ.)~$\uparrow$} \\
    
    \midrule
    \textbf{Ours} (direct, from scratch) & 31.2\% & 28.6\% \\ %
    \textbf{Ours} (gradual, from scratch) & 34.4\% & 32.1\% \\ %

    \midrule
    \textbf{Ours} (direct, finetuned) & 36.7\% & 35.4\% \\ %
    \textbf{Ours} (gradual, finetuned) & \textbf{39.0\%} & \textbf{37.7\%} \\ %
    
  \bottomrule
  \end{tabular}
  }
  \caption{
  \textbf{Ablation study results on ParallelDomain in semantic space.}
  We perform semantic completion of the scene, again similarly to Table~\ref{tab:kubric_abl}. See Figure~\ref{fig:pardom_abl} for qualitative illustrations.
  \vspace{-0.1in}
  }
  \label{tab:pardom_sem_abl}
\end{SCtable}

\section{Choice of camera trajectory}
\label{sec:traject}

Our formulation of the dynamic view synthesis task in Section~\ref{sec:method} is quite general, so it is worth thinking about which specific instantiations of this conceptual framework would be most effective in practice.
Given arbitrary video inputs, our goal is to devise a structured protocol for choosing relative camera trajectories that both maximize the exploitation of knowledge contained within the pretrained SVD representation, as well as enable a detailed understanding of the dynamic scene observed at inference time to the fullest extent possible.
Specifically, we wish to synthesize views that reach as far as the opposite end of the scene, for example, by orbiting the azimuth angle $\phi$ up to 180\textdegree. This is considerably more dramatic than what the state of the art in dynamic view synthesis is typically capable of~\cite{gao2022monocular,cao2023hexplane,li2023dynibar,wu20234dgaussians}, and allows us to reveal large, formerly unseen portions of the surroundings.

However, it turns out that opposing forces are at play. On one hand, we wish to get to the destination camera pose \emph{``as fast as possible''} (because the scene could already be evolving and changing over time as we are watching it). On the other hand, if the output video moves away from the source viewpoint too quickly, we might risk incurring a \emph{distribution misalignment} due to the fact that the image-to-video SVD model predominantly generates videos that start at nearly the exact same spatial perspective as the given image. Moreover,
the camera generally does not move much throughout the video, typically performing only minor panning motions and/or mild rotations.

To resolve this concern, we translate it into three questions: (1) where should the output pose \emph{start}; (2) how fast should it be taught to \emph{move} in-between subsequent frames; and (3) how much does finetuning, \ie borrowing priors from SVD help (or hurt) in each case, versus training an identical network from scratch? We investigate this by running comparative studies on both the Kubric-4D and ParallelDomain-4D datasets.
For each tested scene, we fix a source pose $\mathcal{P}_{{src}}$ and a destination pose $\mathcal{P}_{{dst}}$, following definitions in Section~\ref{sec:task}.
Using $\mathcal{E}_{src,t} = g(\mathcal{P}_{src})$ and $\alpha = \frac{t}{T-1} \in [0,1]$, we define \emph{gradual} and \emph{direct} trajectories as follows:
\begin{align}
    \mathcal{E}_{dst,t} =
\begin{cases}
    g \left( \alpha \mathcal{P}_{dst} + (1-\alpha) \mathcal{P}_{src} \right), & \forall t, \hspace{0.4cm} \text{if gradual} \\
    g \left( \mathcal{P}_{{dst}} \right), & \forall t, \hspace{0.4cm} \text{if direct}
\end{cases}
\end{align}
In other words, \emph{gradual} means that the virtual camera pose corresponding to the output video linearly interpolates (in an intermediate description space, for example spherical coordinates) between $\mathcal{P}_{{src}}$ and $\mathcal{P}_{{dst}}$ from start to end,
whereas \emph{direct} implies that the generated video directly adheres precisely to $\mathcal{P}_{{dst}}$ at every frame without interpolation. For Kubric-4D, \emph{max 90\textdegree} limits the relative horizontal (\ie azimuth) angle variation between input and output to $\pm90\degree$ at training time ($|\Delta \phi| \leq 90\degree$), whereas \emph{max 180\textdegree} effectively allows for synthesizing any 360\textdegree-surround viewpoint of the dynamic scene.

The results are shown in Tables~\ref{tab:kubric_abl},~\ref{tab:pardom_rgb_abl}, and~\ref{tab:pardom_sem_abl}, and Figures~\ref{fig:kubric_abl} and~\ref{fig:pardom_abl}.
From this ablation study, we observe with Kubric-4D that: (1) it is preferable to gradually interpolate from source to destination pose than to immediately jump there (+1.17 dB average PSNR improvement between \emph{direct} and \emph{gradual}); (2) there exists a trade-off between the range of camera transformations the model should be trained for, and how extreme of a rotation one wishes to be able to achieve at most (+0.55 dB between \emph{max 180\textdegree} and \emph{max 90\textdegree}); and (3) it is preferable to start from the SVD checkpoint that had been trained on large-scale video rather than to train from random initialization, though not by a particularly huge margin (+1.34 dB between \emph{scratch} and \emph{finetuned}).

We make consistent findings in the ParallelDomain-4D dataset, where \emph{gradual, finetuned} is the best model.
For Kubric-4D, although \emph{gradual, max 90\textdegree, finetuned} and \emph{gradual, max 180\textdegree, finetuned} are very close, we proceed with the former in all further experiments, described below.

\begin{figure}[tb]
  \centering
  \includegraphics[width=\linewidth]{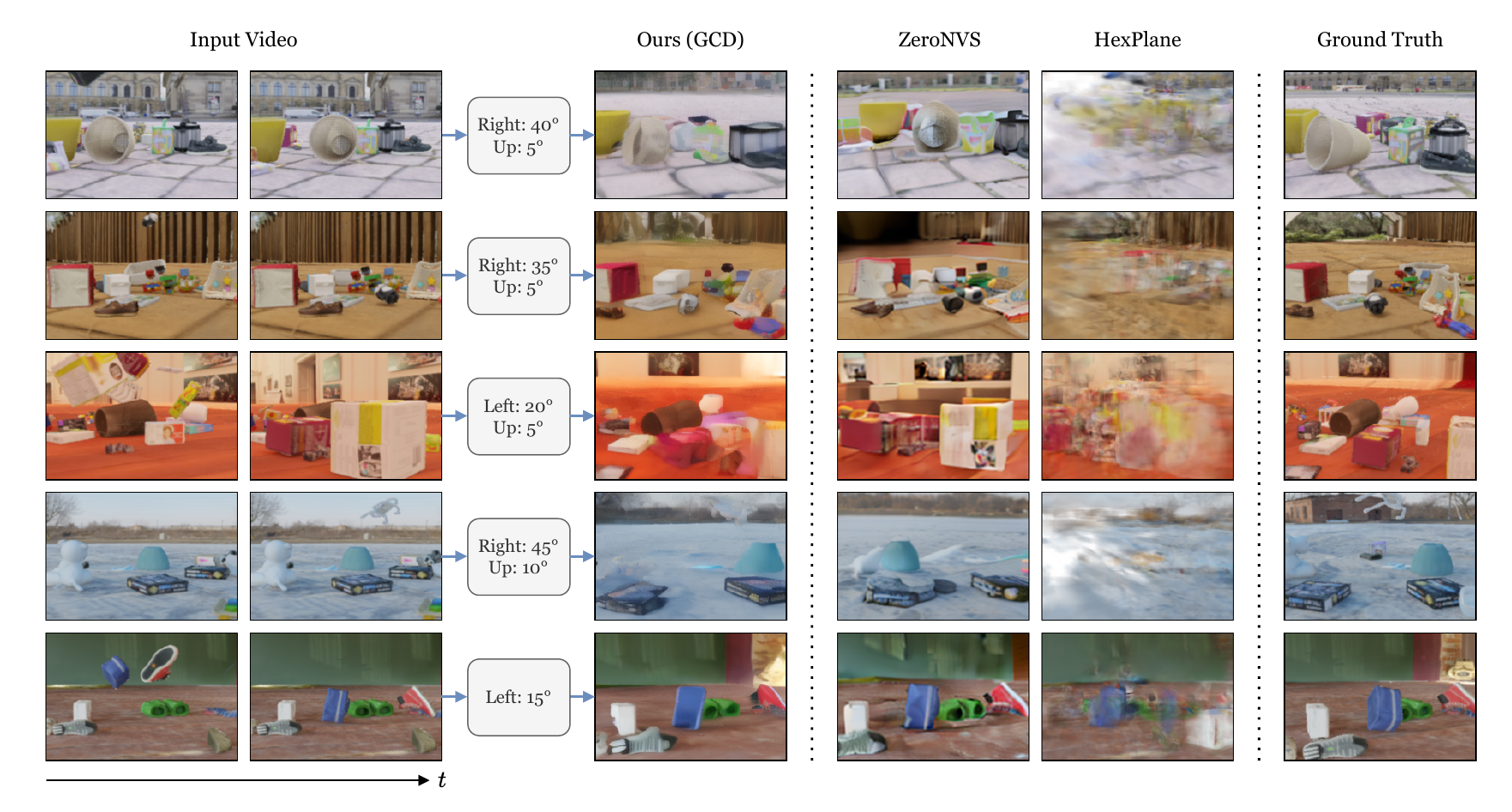}
  \vspace{-0.05in}
  \caption{
  \textbf{Qualitative baseline comparison results for Kubric-4D.}
  We show inputs, predictions, baselines, and ground truths.
  Compared to baselines, our results depict the scene layout and dynamics under the desired novel viewpoints with reasonable accuracy overall and much fewer flickering artefacts.
  \vspace{-0.1in}
  }
  \label{fig:kubric_sota}
\end{figure}

\begin{figure}[tb]
  \centering
  \includegraphics[width=\linewidth]{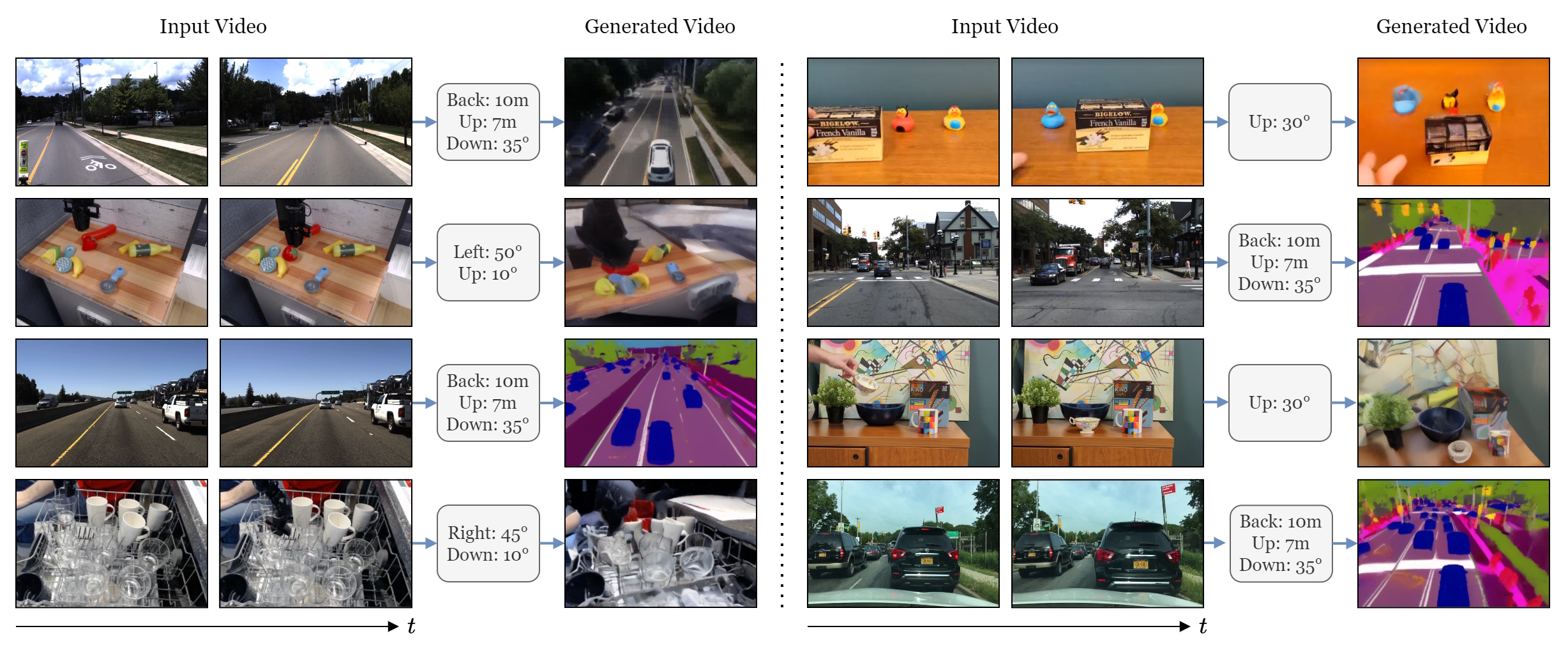}
  \vspace{-0.05in}
  \caption{
  \textbf{Qualitative real-world generalization results.}
  We show inputs and predictions on BridgeData V2 \cite{walke2023bridgedata}, TCOW Rubric \cite{van2023tracking}, TRI-DDAD \cite{packnet}, and Berkeley DeepDrive \cite{yu2020bdd100k}. Despite being trained on synthetic data alone, our approach show surprisingly strong generalization skills to a variety of real-world scenarios. 
  For example, on the top right, where a full occlusion occurs around the middle of the video, our model faithfully predicts both the position and appearance of the invisible duck at the last frame, demonstrating object permanence capabilities.
  \vspace{-0.1in}
  }
  \label{fig:realworld}
\end{figure}

\begin{table}[tb]
  \centering
  \scalebox{0.9}{
  \begin{tabular}{@{}l|rrrrr@{}}
    \toprule
    Method & \thead{PSNR\\ (all)~$\uparrow$} & \thead{SSIM\\ (all)~$\uparrow$} & \thead{LPIPS\\ (all)~$\downarrow$} & \thead{PSNR\\ (occ.)~$\uparrow$} & \thead{SSIM\\ (occ.)~$\uparrow$} \\
    
    \midrule
    HexPlane~\cite{cao2023hexplane} & 15.38 & 0.428 & 0.568 & 14.71 & 0.428 \\
    4D-GS~\cite{wu20234dgaussians} & 14.92 & 0.388 & 0.584 & 14.55 & 0.392 \\
    DynIBaR~\cite{li2023dynibar} & 12.86 & 0.356 & 0.646 & 12.78 & 0.358 \\
    
    \midrule
    Vanilla SVD~\cite{blattmann2023stable} & 13.85 & 0.312 & 0.556 & 13.66 & 0.326 \\
    ZeroNVS~\cite{sargent2023zeronvs} & 15.68 & 0.396 & 0.508 & 14.18 & 0.368 \\

    \midrule
    \textbf{Ours} & \textbf{20.30} & \textbf{0.587} & \textbf{0.408} & \textbf{18.60} & \textbf{0.527} \\ %
    
    \midrule
    Reproject RGB-D\tss{*} & 12.51 & 0.537 & 0.416 & - & - \\ %
  
  \bottomrule
  \end{tabular}
  }
  \vspace{0.07in}
  \caption{
  \textbf{Baseline comparison results on Kubric-4D.} We evaluate gradual dynamic view synthesis models on all 13 output frames, and with a single RGB video as input.
  We significantly outperform all baselines for both visible and occluded pixels.
  {\smaller \tss{*}Uses privileged information, \ie can access the ground truth depth map from the input viewpoint.}
  \vspace{-0.2in}
  }
  \label{tab:kubric_sota}
\end{table}

\begin{table}[tb]
  \centering
  \scalebox{0.9}{
  \begin{tabular}{@{}l|rrrrr@{}}
    \toprule
    Method & \thead{PSNR\\ (all)~$\uparrow$} & \thead{SSIM\\ (all)~$\uparrow$} & \thead{LPIPS\\ (all)~$\downarrow$} & \thead{PSNR\\ (occ.)~$\uparrow$} & \thead{SSIM\\ (occ.)~$\uparrow$} \\
    
    \midrule
    Vanilla SVD~\cite{blattmann2023stable} & 12.88 & 0.400 & 0.658 & 13.96 & 0.466 \\
    ZeroNVS~\cite{sargent2023zeronvs} & 18.88 & 0.490 & 0.555 & 19.29 & 0.552 \\
    
    \midrule
    \textbf{Ours} & \textbf{25.04} & \textbf{0.731} & \textbf{0.358} & \textbf{24.70} & \textbf{0.733} \\ %
    
    \midrule
    Reproject RGB-D\tss{*} & 17.66 & 0.459 & 0.441 & - & - \\ %
  \bottomrule
  \end{tabular}
  }
  \vspace{0.07in}
  \caption{
  \textbf{Baseline comparison results on ParallelDomain in RGB space.}
  We perform visual scene completion, and evaluate gradual dynamic view synthesis on all 13 output frames, and with a single RGB video as input.
  We significantly outperform all baselines for both visible and occluded pixels.
  {\smaller \tss{*}Uses privileged information, \ie can access the ground truth depth map from the input viewpoint.}
  \vspace{-0.2in}
  }
  \label{tab:pardom_rgb_sota}
\end{table}

\begin{SCtable}
  \centering
  \scalebox{0.9}{
  \begin{tabular}{@{}l|rr@{}}
    \toprule
    Method & \thead{mIoU\\ (all)~$\uparrow$} & \thead{mIoU\\ (occ.)~$\uparrow$} \\
    
    \midrule
    \textbf{Ours} & \textbf{43.4}\% & \textbf{38.2}\% \\ %
    
    \midrule
    Reproject Sem-D\tss{*} & 37.3\% & - \\ %
  \bottomrule
  \end{tabular}
  }
  \caption{
  \textbf{Baseline comparison results on ParallelDomain in semantic space.}
  We perform semantic completion of the scene, still based on a single RGB video as input.
  {\smaller \tss{*}Uses privileged information, \ie can access the ground truth depth map \emph{and} ground truth semantic category of all input pixels.}
  \vspace{-0.1in}
  }
  \label{tab:pardom_sem_sota}
\end{SCtable}

\section{Experiments}
\label{sec:exp}

In this section, we evaluate our monocular dynamic novel view synthesis framework.
We report numerical results on the test splits of our two in-domain datasets (Kubric-4D and ParallelDomain-4D), comparing against several state-of-the-art baselines, but additionally showcase promising qualitative results on in-the-wild videos from various domains.
For more results as well as animated visualizations, please see \href{https://gcd.cs.columbia.edu/}{gcd.cs.columbia.edu}.

\subsection{Implementation details}

\paragraph{Training.}
We adopt the SVD variant that predicts $T=14$ frames, but due to computational constraints, we downscale the input and output resolution to $W \times H = 384 \times 256$.
This allows us to scale the batch size up to 56 when training with Kubric-4D on 7x A100 GPUs with 80 GB VRAM each. 
We finetune all models for 10k iterations using the Adam optimizer, which takes roughly 3 days.
On ParallelDomain-4D, we instead finetune models for 13k iterations with an effective batch size of 24 through a gradient accumulation factor of 4 on 3x A6000 GPUs with 48 GB VRAM each, which also takes roughly 3 days.
The network $\epsilon$ does not predict noise directly, instead adopting v-parameterization for preconditioning~\cite{salimans2022progressive}.

\paragraph{Inference.}
We generate conditional samples from the resulting diffusion model by running the EDM sampler~\cite{karras2022elucidating} for 25 steps. SVD originally employs classifier-free guidance~\cite{ho2022classifier} at test time with a guidance strength $w$ that linearly increases as a function of the video frame index (not the diffusion timestep) from start to end within the range $[1,2.5]$ by default~\cite{blattmann2023stable}, but we found better performance by adjusting this range to $[1, 1.5]$ instead. Producing one output video takes roughly 10 seconds.

\paragraph{Evaluation metrics.}
Following related work in novel view synthesis~\cite{mildenhall2020nerf,kerbl3Dgaussians,sargent2023zeronvs,liu2023zero,li2023dynibar,wang2024diffusion}, for predictions in RGB space, we evaluate PSNR, SSIM, and LPIPS scores and average the results across both video frames and test examples.
For semantic category predictions, following conventions in semantic segmentation~\cite{cordts2016cityscapes,xie2021segformer,strudel2021segmenter,zheng2021rethinking}, we first calculate the average Intersection over Union (IoU) per category over the whole ParallelDomain test set, and then report the mean IoU (mIoU) across the 10 most common categories.

Based on the ground truth depth information from the input viewpoint, it is also possible to determine which pixels in the target viewpoint are visible or hidden. In addition to the regular metrics (``all''), we therefore spatially mask the videos to determine metrics for occluded regions only (``occ.''), which the model essentially has to inpaint.

Even though our model accepts and predicts the same number of frames ($T=14$), the first output frame for the \emph{gradual} camera trajectory models (described below) is spatially aligned with the first input frame. This implies that it could in principle be solved by copying its pixels (except if the task involves switching to another modality, for example semantic category prediction), so we exclude the first frame from the evaluation to avoid inflating the metrics, instead averaging only over the last $T-1=13$ frames, which correspond to different extrinsics.

\subsection{Baselines}

We compare our final models against the state-of-the-art dynamic view synthesis methods including \textit{HexPlane}~\cite{cao2023hexplane}, \textit{4D-GS}~\cite{wu20234dgaussians} and \textit{DynIBaR}~\cite{li2023dynibar}, which all perform per-scene optimization.
While these baselines are capable of handling videos with higher resolutions than ours, they are typically limited to much smaller camera angle changes in the one- or low-number-of-views regime, and inference runtimes are many orders of magnitude larger (\eg hours vs. seconds).

In addition, we compare to two pretrained diffusion models \textit{Vanilla SVD}~\cite{blattmann2023stable} and \textit{ZeroNVS}~\cite{sargent2023zeronvs} by adapting them for our task.
For Vanilla SVD, we run the original SVD model to generate videos based on the first input frame, without any changes or finetuning.
For ZeroNVS, which can generate novel views of scenes based on a single image, we run it for all the input frames independently to obtain the output video.

Finally, we compare to a simple geometric baseline (\textit{Reproject RGB-D} and \textit{Reproject Sem-D}), where we reproject pixels from input frames to target viewpoints using the ground truth depth maps, switching to the appropriate modality as needed.
Here, the goal is to study how much information is contained within the input video itself, if precise per-pixel depth values were fully known (which is often not the case).

All methods observe the same monocular input video, and are evaluated on the exact same set of randomly sampled output camera trajectories for fairness.

\subsection{Results}

We report quantitative results in Tables~\ref{tab:kubric_sota},~\ref{tab:pardom_rgb_sota}, and~\ref{tab:pardom_sem_sota}, and show qualitative results in Figures~\ref{fig:kubric_sota} and~\ref{fig:pardom_abl}.
On both datasets, our model outperforms baseline methods by a large margin.
Per-scene optimization methods (e.g., HexPlane) fail to reconstruct the 4D scene representation from a single input view, and thus the rendered videos from novel viewpoints have severe artifacts.
Vanilla SVD is able to generate smooth videos but fails to follow the desired camera trajectories, and does not incorporate content from later frames.
ZeroNVS can synthesize plausible individual frames from specified viewpoints, but the resulting videos are not temporally coherent and do not respect the scene dynamics.

In contrast, our model mostly generates plausible videos that accurately depict the complex scene geometry and motion under the desired novel viewpoint transformations.
We remark that the results are not perfect, as the correspondence of objects between the input and generated output videos is not always very clear, and some rigid objects tend to erroneously deform. However, the nature of the task is extremely challenging, and we expect the potential visual quality and consistency of our framework to only improve over time, \eg when combined with better pretrained representations, more compute, and more data.

Apart from the evaluation on in-domain datasets, we also showcase promising results on real-world in-the-wild videos. As shown in Figure~\ref{fig:realworld}, our model sometimes generalizes quite well to various domains including driving environments, daily indoor videos, and robotic manipulation scenes.

\section{Discussion}
\label{sec:disc}
In this paper, we present a framework for dynamic novel view synthesis from a monocular video by finetuning a large-scale pretrained video diffusion model~\cite{blattmann2023stable} on high-quality synthetic data. 
While we show promising results on real-world in-the-wild videos, our model still struggles on significantly out-of-distribution examples, \eg videos with moving humans.
Nevertheless, we believe this work delivers meaningful progress in terms of gaining a rich, detailed understanding of 4D scenes, and takes a solid first step towards enabling zero-shot dynamic view synthesis from a monocular video.

{\small
\textbf{Acknowledgements:}
This research is based on work partially supported by the NSF CAREER Award \#2046910 and the NSF Center for Smart Streetscapes (CS3) under NSF Cooperative Agreement No.\ EEC-2133516.
The views and conclusions contained herein are those of the authors and should not be interpreted as necessarily representing the official policies, either expressed or implied, of the sponsors.
}

\bibliographystyle{splncs04}
\bibliography{_main}

\renewcommand\thesection{\Alph{section}}
\renewcommand\thesubsection{\thesection.\arabic{subsection}}

\newpage
{
\centering
\Large
\textbf{Generative Camera Dolly: Extreme Monocular Dynamic Novel View Synthesis} \\
\vspace{0.5em}Supplementary Material \\
\vspace{1.0em}
}
\appendix

\section{Overview}

The appendix is structured as follows: in Section~\ref{sec:morequal}, we analyze what the equivalent number of source views given to HexPlane would have to be to match our method's performance, as well as our model's metrics as a function of the geometric ``difficulty'' of the camera controls. In Section~\ref{sec:impl}, we elaborate on implementation details in terms of the model architecture, how training is done, how datasets are processed, how evaluations are performed, and how the baselines are adapted. In Section~\ref{sec:fail}, we discuss failure cases.
To view video visualizations of extra qualitative results,
we recommend viewing \href{https://gcd.cs.columbia.edu/}{gcd.cs.columbia.edu} in a modern web browser.

\begin{SCfigure}[][b!]
\centering
  \includegraphics[width=0.5\linewidth]{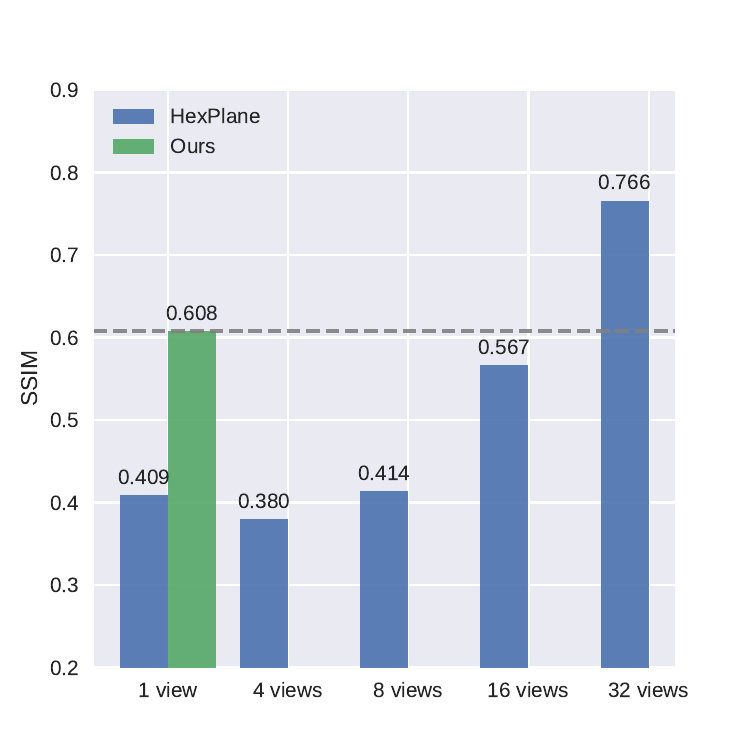}
  \caption{
  \textbf{Comparative study over number of views.}
  We plot the SSIM over the test set as a function of the number of input views that HexPlane uses for training. The numbers are averaged over 20 scenes.
  }
  \label{fig:dnerf_chart}
\end{SCfigure}

\begin{SCfigure}[][t!]
  \centering
  \includegraphics[width=0.5\linewidth]{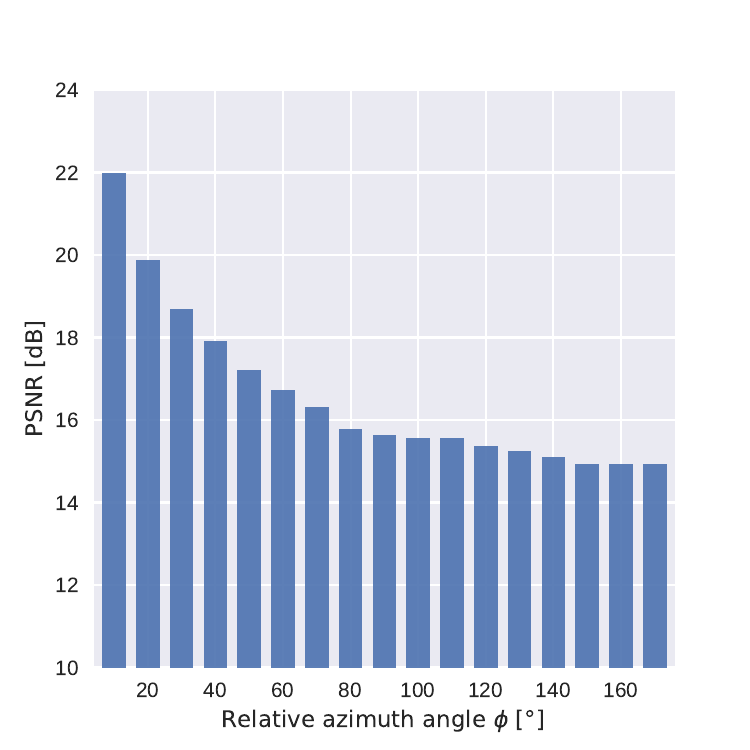}
  \caption{
  \textbf{Comparative study over camera rotation magnitude in Kubric-4D.}
   Note that PSNR is measured at the last output frame, because only then the desired horizontal azimuth angle has been reached. We conclude that the main difficulty in performing dynamic view synthesis comes from handling roughly the first 80 degrees, after which the performance stays mostly flat.
  }
  \label{fig:error_chart}
\end{SCfigure}

\section{More quantitative evaluations}
\label{sec:morequal}

\subsection{Comparison to multi-view methods}
\label{sec:kview}

Our method is able to synthesize novel views of a dynamic scene from just a single-viewpoint input video.
One other hand, the results from per-scene optimization methods (e.g., HexPlane~\cite{cao2023hexplane}) get better with an increasing number of input views.
A natural question is that how many input views are needed for those methods in order to obtain similar performance as compared to ours from a single view.
We try to answer this question by training HexPlane per scene with $K$ training views (i.e., $K$ input videos), with $K \in \{1,4,8,16,32\}$.
As shown in Figure~\ref{fig:dnerf_chart}, our results (from a single input view) give rise to even better quality than HexPlane's results from $16$ input views.

\subsection{Error as a function of rotation angle}
\label{sec:error}

In Figure~\ref{fig:error_chart}, we plot the average PSNR over the test set as a function of how significantly the final destination (target) camera pose differs from the source (input) camera pose.
Specifically, we evaluate the Kubric-4D (\emph{gradual, max 180\textdegree, finetuned}) model on a sequence of horizontal rotations to the right of varying amounts between 0\textdegree\ and 180\textdegree.
The elevation angle $\theta$ is held constant at 10\textdegree, to encourage obstructed objects from the input view, and the radius $r$ at 15m.

\begin{SCfigure}[][b!]
  \centering
  \includegraphics[width=0.4\linewidth]{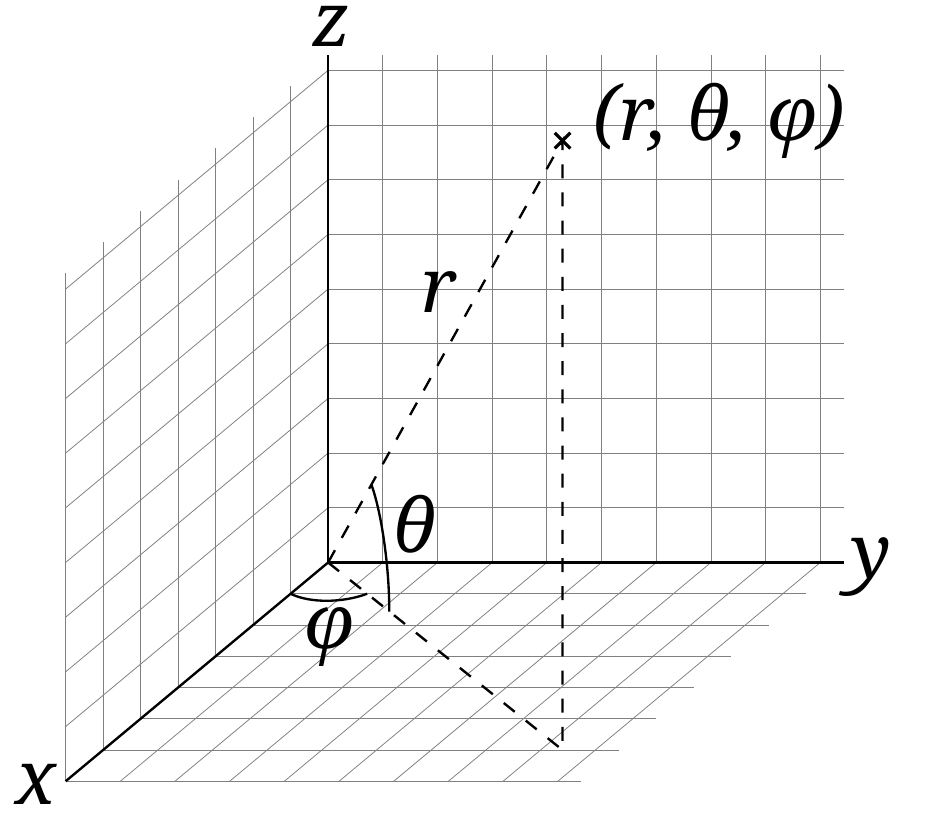}
  \vspace{-0.05in}
  \caption{
  \textbf{Spherical coordinate system.} Models trained on Kubric-4D accept an azimuth $\phi$, elevation $\theta$, and radius $r$ as input to condition the video generation process. (Illustration adapted from~\cite{wikispher}.)
  }
  \label{fig:coord}
\end{SCfigure}

\section{Implementation details}
\label{sec:impl}

\subsection{Coordinate system}

We use a spherical coordinate system, where $(\phi, \theta, r)$ represents the azimuth angle, elevation angle, and radial distance respectively. Note that as shown in Figure~\ref{fig:coord}, $\theta$ is the \emph{elevation} angle as measured starting from the XY-plane, which is not the same as the \emph{inclination} angle as measured starting from the Z-axis. %

\subsection{Architecture}

\begin{figure}[tb]
  \centering
  \includegraphics[width=\linewidth]{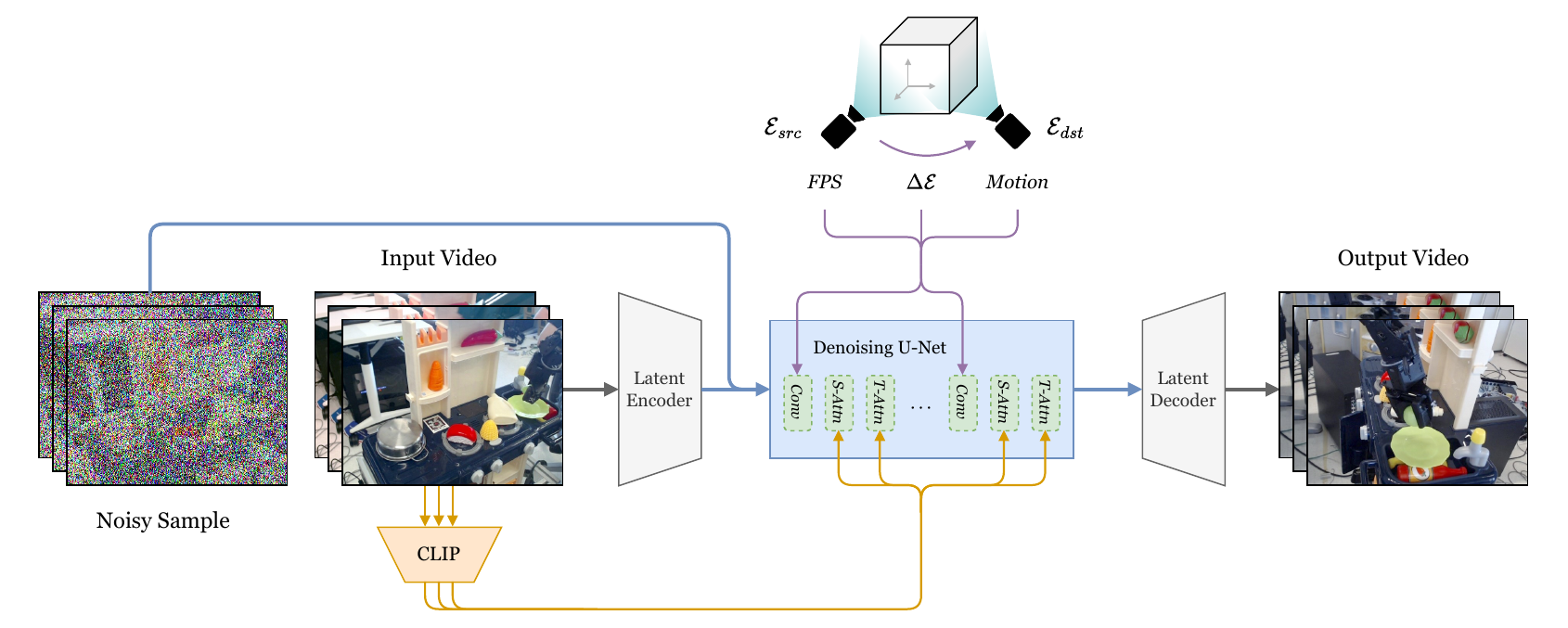}
  \vspace{-0.05in}
  \caption{
  \textbf{Network architecture.}
  Our model performs diffusion in latent space~\cite{rombach2022high,ling2023align}. The input video is encoded by a KL-VAE, and then channel-concatenated with the noisy sample. At training time, the output video is estimated and supervised; at inference time, multiple denoising steps are performed. In both cases, per-frame CLIP embeddings condition the U-Net by means of cross-attention, and and other relevant pieces of information (frame rate, desired camera pose transformation, and motion value) condition the U-Net by adding their embeddings onto the feature vectors in-between convolutions.
  \vspace{-0.1in}
  }
  \label{fig:arch}
\end{figure}

Figure~\ref{fig:arch} describes the model architecture in more detail. It is based on SVD~\cite{blattmann2023stable}, which in turn is based on Video LDM~\cite{blattmann2023align}, modified for camera pose conditioning.
The $T=14$ CLIP embeddings are fed to the network via multiple spatial (S-Attn) and temporal (T-Attn) cross-attention blocks throughout the network.
Separately, the \emph{micro-conditioning} mechanism takes place to pass the embeddings of the diffusion timestep, frame rate, camera transformation, motion bucket value, and conditioning augmentation strength to the network by summing it together with feature channels at various residual blocks placed throughout the network, with additional linear projections in-between to accommodate varying embedding sizes.
Concretely, assuming the camera always looks at the same location in 3D space for simplicity,\footnote{This is $(0, 0, 1)$, \ie 1m above the center of the ground plane, in Kubric-4D.} the relative extrinsics matrix $\Delta \mathcal{E}$ is parameterized and given as $(\Delta \phi, \Delta \theta, \Delta r)$. The angles are subsequently encoded with Fourier positional encoding before being embedded through an MLP.
Note that the input camera poses are not required to be known -- only the desired relative transformation should be given.

\subsection{Data and training}

In Kubric-4D, pairs of input and output video clips are always temporally synchronized, but with $T=14$ frame indices sampled randomly within the 60 available frames from the dataset. The original FPS is 24, and since the frame stride is randomly uniformly sampled among $\{1,2,3,4\}$, the actual FPS when finetuning therefore belongs to $\{6,8,12,24\}$. In ParallelDomain-4D, each scene has 50 frames available at 10 FPS, from which we randomly subsample clips but only at a frame stride in $\{1,2\}$ determined by a coin flip, which implies an FPS value in $\{5,10\}$.

In Kubric-4D, the camera pose $\mathcal{P}=(\phi, \theta, r)$ respects the following bounds (both across time and across input/output) with respect to the spherical coordinate system: azimuth angle $\phi_{1 \ldots T} \in [0\degree, 360\degree]$, elevation angle $\theta_{1 \ldots T} \in [0\degree, 50\degree]$, radial distance $r_{1 \ldots T} \in [12, 18]$.\footnote{
Since the dataset is synthetic and the radius $r$ does not have an inherent meaning, it is worth nothing that the average diameter of an object is 1.88m, and that all objects are randomly spawned within these bounds in Euclidean coordinates: $x \in [-7, 7], y \in [-7, 7], z \in [0, 7]$ (where Z is up).}
The target camera pose transformation for the default model (\emph{max 90\textdegree}) has a limited maximum transformation ``strength'' in the sense that from start to end, the azimuth, elevation, and radius all vary within the following bounds: $|\Delta \phi| \leq 90\degree, |\Delta \theta| \leq 30\degree, |\Delta r| \leq 3$. The horizontal field of view is 53.1\textdegree\ everywhere.

For the more extreme view synthesis variant (\emph{max 180\textdegree}), the bounds are: $\phi_{1 \ldots T} \in [0\degree, 360\degree], \theta_{1 \ldots T} \in [0\degree, 90\degree], r_{1 \ldots T} \in [12, 18], |\Delta \phi| \leq 180\degree, |\Delta \theta| \leq 60\degree, |\Delta r| \leq 3$.

The trajectories are typically uniformly sampled, except for the elevation angle $\theta$; in this case, uniform sampling for the starting point happens in terms of $\sin \theta$ instead of the angle $\theta$ directly. This is done in order to ensure an equal spread over (\ie a uniform distribution on the surface of) the (relevant subset of the) unit sphere. The input camera extrinsics $\mathcal{E}_{src,t}$ is static, and the output camera extrinsics $\mathcal{E}_{dst,t}$ interpolates linearly over time in pose description space, \ie in spherical coordinates with $\alpha=\frac{t}{T-1}$.

In ParallelDomain-4D, the source viewpoint is a forward-facing camera mounted on the virtual ego car at a fixed position of $(1.6, 0, 1.55)$ in 3D world space, where X points forward and Z points up. For simplicity, the camera pose is not controllable in the experiments described in our paper -- instead, the destination viewpoint is fixed at $(-8, 0, 8)$, looking forward and down at $(5.6, 0, 1.55)$. To maximize the temporal smoothness of the generated video, the camera trajectory is interpolated in Euclidean space, not linearly but rather according to a sine wave function, \ie following $\alpha = \frac{ 1 - \cos \left( \frac{t}{\pi (T - 1)} \right) }{2}$, assuming $t$ increases step-wise from 0 to $T-1$. The horizontal field of view is 85\textdegree\ everywhere.

Early on in our experiments, we observed that synchronizing the motion bucket value, which conditions the model, with the strength of the camera transformation leads to better performance. Therefore, for Kubric-4D, we linearly scale this value along with the magnitude of the relative camera rotation (specifically, the $\mathcal{L}_2$ norm of $(\Delta \phi, \Delta \theta)$) where the minimum value corresponds to 0 and the maximum value corresponds to 255. This indication of camera motion hints the model that it should generate a video with a high degree of optical flow when the relative angles are high and vice versa.

We keep conditioning augmentation~\cite{blattmann2023align} enabled with a noise strength of 0.02.

\subsection{Loss}

We apply a focal $\mathcal L_2$ loss function between the estimated and ground truth latent feature maps, which focuses on the top fraction of embeddings incurring the biggest mismatch. This fraction linearly decreases from $100\%$ to $10\%$ in the first 5000 iterations, and then remains constant at $10\%$. In addition, for semantic completion in ParallelDomain-4D, we weight the categories involving vehicles (\ie \textit{Bus, Car, Caravan/RV, ConstructionVehicle, Bicycle, Motorcycle, OwnCar, Truck, WheeledSlow}) and people (\ie \textit{Animal, Bicyclist, Motorcyclist, OtherRider, Pedestrian}) to be respectively $3\times$ and $7\times$ as important as other categories, by multiplying the loss values at the corresponding spatial positions with the appropriate scaling factor before averaging. We observe that this strategy tends to reduce false negative prediction rates, especially for visually smaller objects occupying fewer pixels.

\subsection{Evaluation}

For each dataset separately, all models and all variants are evaluated on the same test split. For each scene, we randomly sample a subclip within the available video with $T=14$ frames and a variable frame rate chosen within the same range as during training time. Then, for Kubric-4D, four different target camera poses (with angles up to azimuth $\pm90\degree$ for Kubric-4D) are randomly sampled once.
To encourage difficult input videos with higher than average degrees of occlusion, we set the starting elevation angle to be always $\theta_1=5\degree$, but all other angles are chosen randomly within the same ranges as during training.
These randomization parameters at test time are only chosen once and then fixed across all evaluation experiments.
We let probabilistic (\ie diffusion) models (Ours, Vanilla SVD, ZeroNVS) generate four samples for each of these trajectories, averaging results, but the other methods (HexPlane, 4D-GS, DynIBaR) are only executed once for each scene and for each set of output camera angles.

\subsection{Baselines}

\paragraph{Vanilla SVD~\cite{blattmann2023stable}.} Since Stable Video Diffusion's last training stage involved finetuning at a resolution of $1024 \times 576$, and changing the resolution at test time gives rise to artefacts, it is probably optimal to evaluate the model at its original resolution. We center crop and resize all input images and target videos as needed to $1024 \times 576$ when evaluating this baseline. We keep the motion bucket at its default value of 127.

\paragraph{ZeroNVS~\cite{sargent2023zeronvs}.}
Like Zero-1-to-3~\cite{liu2023zero}, ZeroNVS was trained only on square images of resolution $256 \times 256$.
Similarly to Vanilla SVD, we center crop and resize all input and ground truth frames accordingly.
Moreover, since ZeroNVS learns a scale-invariant means of transforming camera poses in a way that depends on estimated depth maps, the translation component of the relative camera extrinsics matrix $\mathcal{E}_{src}^{-1} \cdot \mathcal{E}_{dst}$ fed to the model can incur variable meanings with respect to absolute 3D space depending on the observed scene. A scale parameter is hence tuned visually for each video separately until the output qualitatively aligns with the ground truth.

\begin{figure}[tb]
  \centering
  \includegraphics[width=\linewidth]{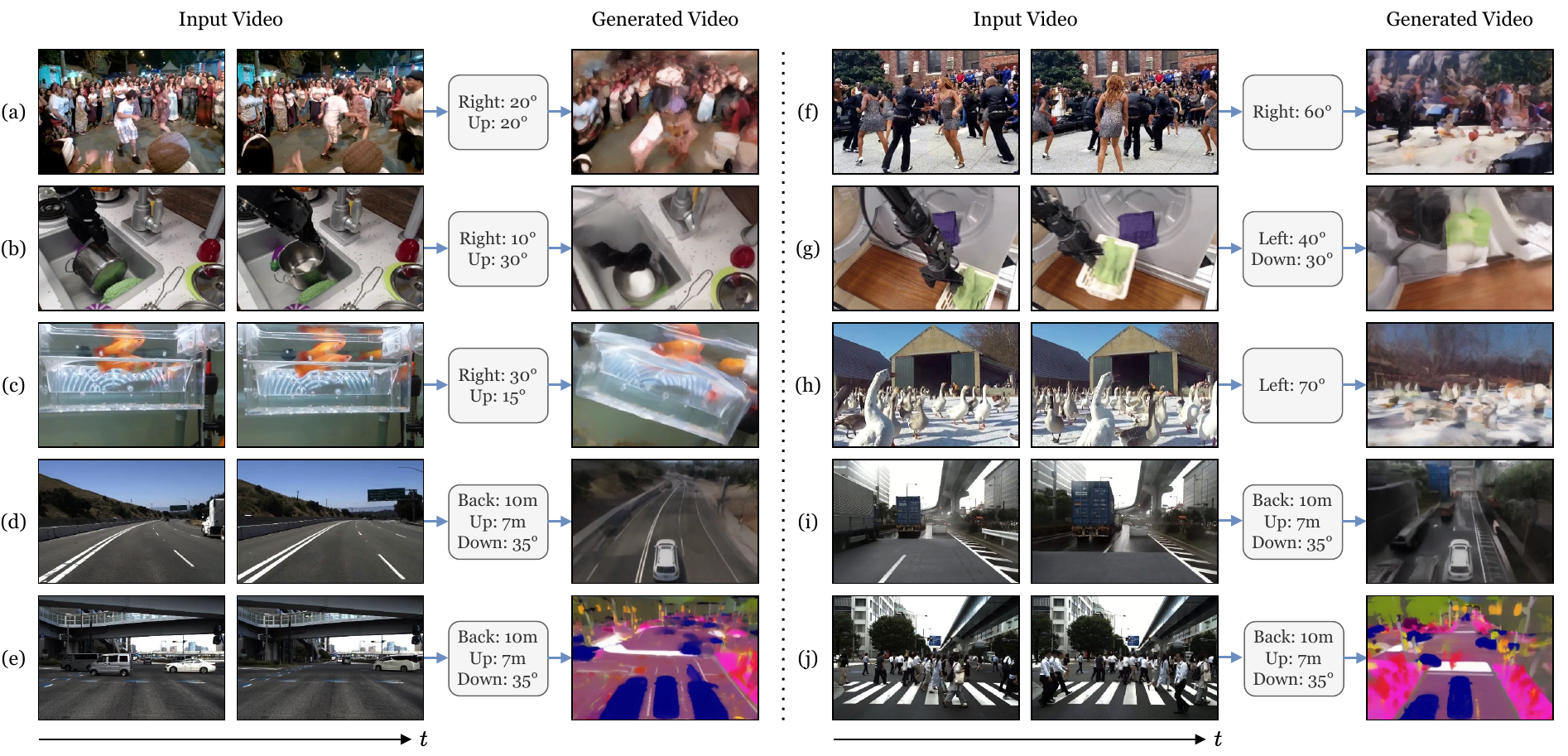}
  \vspace{-0.05in}
  \caption{
  \textbf{Failure cases.}
  We show inputs and predictions of real-world examples. Since deformable objects are not present in our Kubric-4D finetuning set, our model occasionally struggles with reconstructing their shape, appearance, and motion correctly. This can sometimes lead to objects becoming vague or blending in with each other. Similarly, videos in the bottom two rows are possibly related to them bordering on being out-of-distribution with respect to ParallelDomain-4D.
  \vspace{-0.1in}
  }
  \label{fig:fail}
\end{figure}

\section{Failure cases}
\label{sec:fail}

Our model exhibits strong performance in many cases, but also fails to accurately generalize to some real-world videos, especially those involving humans, animals, or deformable objects.
In Figure~\ref{fig:fail}, we show representative failure cases.
In (a), while the general layout is somewhat preserved, the people themselves become blurry.
In (b), the robot arm gets cut off when performing view synthesis from the top, presumably because Kubric-4D does not contain robots or robotic motion patterns.
In (c), the model appears to be confused as to what the initial camera pose is, and interprets it as a top-down rather than a sideways perspective of an aquarium, which leads to a roll effect when rotating the azimuth.
In (d), the highway sign gets missed.
In (e), the overpasses are not reconstructed, which seems to cause blurriness in the rest of the prediction.
In (f), (g), and (h), both shape and dynamics are not well-respected.
In (i), the perceived depth of the large blue truck is wrong.
In (j), there are an unusually large amount of pedestrians crossing the street, which the model groups into ``cars''.

\end{document}